\documentclass[sigconf, natbib=false]{acmart}

\AtBeginDocument{%
  }

\RequirePackage[
  datamodel=acmdatamodel,
  style=acmnumeric,
  ]{biblatex}

\usepackage{booktabs}
\usepackage{array}
\usepackage{tcolorbox}
\usepackage{amsmath}
\usepackage{multirow}
\usepackage{multicol}
\usepackage{makecell}
\usepackage{enumitem}
\usepackage{hyperref}
\hypersetup{
    colorlinks=true,
    linkcolor=blue,
    filecolor=magenta,      
    urlcolor=cyan,
}

\newcommand{\algname}{\texttt{Ext2Gen}} 
\newcolumntype{L}[1]{>{\raggedright\let\newline\\\arraybackslash\hspace{0pt}}m{#1}}
\newcolumntype{X}[1]{>{\centering\let\newline\\\arraybackslash\hspace{0pt}}p{#1}}

\begin{document}

\title{Aligning Extraction and Generation for Robust Retrieval-Augmented Generation}

\author{Hwanjun Song}
\email{songhwanjun@kaist.ac.kr}
\affiliation{%
  \institution{KAIST}
  \country{Republic of Korea}
}
\authornote{Hwanjun Song is the first and corresponding author.}

\author{Jeonghwan Choi}
\email{hwani.choi@kaist.ac.kr}
\affiliation{%
  \institution{KAIST}
  \country{Republic of Korea}
}

\author{Minseok Kim}
\email{minseokkim0630@gmail.com}
\affiliation{%
  \institution{Meta}
  \country{United States}
}


\begin{abstract}
Retrieval-augmented generation (RAG) enhances LLMs with external knowledge, yet generation remains vulnerable to retrieval-induced noise and uncertain placement of relevant chunks, often causing hallucinations. We present {Ext2Gen}, an extract-then-generate framework that strengthens LLMs via joint evidence selection and answer generation, dynamically identifying query-relevant content while suppressing noise, thereby removing the need for any independent pre-generation compression module. Optimized through preference alignment with well-curated pairwise feedback, {Ext2Gen} produces accurate and faithful answers even under noisy or imprecise retrieval. Experiments demonstrate that it substantially enhances the robustness of the generation backbone and yields greater performance gains than methods relying on independent compression models (\emph{e.g.}, Recomp, CompAct, EXIT). It further benefits from improved retrieval techniques such as query rewriting, underscoring that generation-side enhancements address limitations that retrieval alone cannot overcome. The trained model is available at \href{https://huggingface.co/DISLab/Ext2Gen-8B-R2}{\color{blue}https://huggingface.co/DISLab/Ext2Gen-8B-R2}.
\end{abstract}




\maketitle

\section{Introduction}
\label{sec:intro}

Retrieval-augmented generation (RAG) has proven its effectiveness in reducing hallucinations in large language models (LLMs), when their knowledge is incomplete, outdated, or lacks sufficient detail to accurately address specific queries \cite{ fan2024survey, rau2024context, wu2025lighter}. A critical aspect of RAG is the "retrieval" process, which involves identifying and selecting relevant text chunks. The quality of these retrieved chunks plays a pivotal role in the overall performance of RAG, as they form the basis for generating factual and contextually relevant answers aligned with the query intent \cite{asaiself, wang2023query2doc, zhang2024mugi}. 

In this regard, most recent works have primarily focused on improving retrieval accuracy to increase the likelihood of relevant chunks being included in the Top-$k$ search results, such as query rewriting \cite{wang2023query2doc, zhang2024mugi}, re-ranking \cite{hwang2024dslr, reddy2024first}, and self-critique \cite{asaiself, li2024you}. These methods work by expanding contextual information to the query, re-scoring retrieved chunks to prioritize relevance, and validating the chunks against the query to ensure consistency.

Despite advancements in retrieval accuracy, bottlenecks persist in the generation process due to two key challenges. First, the \emph{uncertain placement} of relevant chunks often leads to their unpredictable positioning within the retrieved list. This poses a significant challenge for generation, as LLMs are highly sensitive to context order. When relevant information appears in the middle, it may be forgotten due to the {lost-in-the-middle} phenomenon \cite{liu2024lost}. Second, generation is further hindered by \emph{information noisiness}, where irrelevant chunks are included to varying degrees, distracting the model and diluting its focus \cite{cuconasu2024power}. This issue gets more severe as more chunks are retrieved. Although retrieval recall improves, the additional noise makes it harder for the model to identify truly relevant information.
These challenges are particularly pronounced in RAG, where smaller LLMs, more susceptible to noise-related vulnerabilities, are often used for generation (see Section \ref{sec:exp1_main}).


In this paper, we therefore go beyond accurate retrieval to emphasize \emph{robust generation} that remains resilient to information forgetting and noisiness. Unlike prior work that primarily focuses on improving retrieval \cite{choi2025word2passage, li2024you} or maintaining an independent content compression model \cite{hwang2025exit, xu2023recomp, yoon2024compact}, our approach directly enhances the robustness of the generation model itself.
To this end, we propose a novel training framework named \algname{} (Extract-then-Generate), designed to enhance the robustness of any LLM backbone. The core idea is to train the model to first extract query-relevant sentences from \emph{noisy} retrieved chunks and then refine the extracted content to produce a precise answer. This extraction step functions as evidence reasoning \cite{chu2023survey, wei2022chain}, guiding the model toward a reliable final response. Crucially, we move beyond prompt engineering by training the model to identify relevant content and suppress noise, making the LLM inherently more robust.

\begin{figure*}[t!]
\begin{center}
\includegraphics[width=17.2cm]{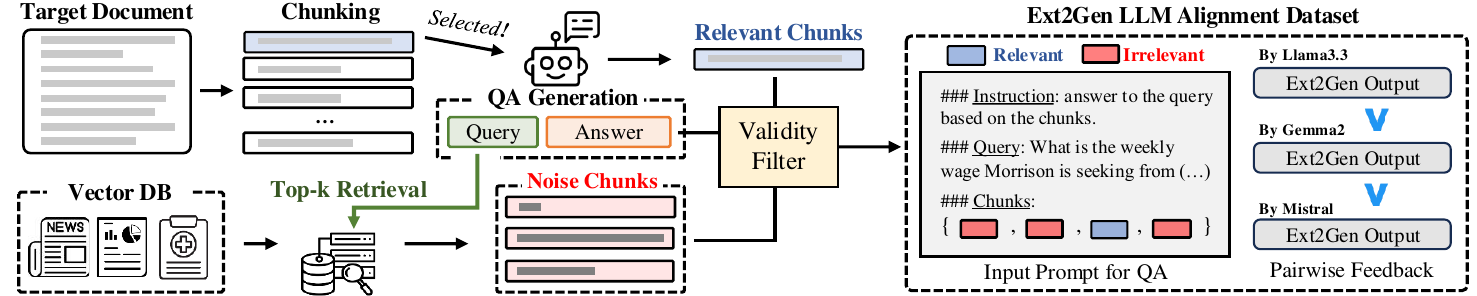}
\end{center}
\vspace*{-0.30cm}
\caption{Overview of Ext2Gen. We simulate noisy RAG inputs by mixing relevant and irrelevant chunks with LLM-generated queries. Multiple LLMs generate answers, and pairwise feedback is derived to train the LLM backbone for robust generation.}
\label{fig:overview}
\vspace*{-0.3cm}
\end{figure*}

To systematically achieve this robustness, we frame the two challenges in generation as an \emph{alignment} problem \cite{wang2024comprehensive, zhou2024lima}, where a discrepancy exists between the model’s desired capability and its actual behavior. Ideally, the model should accurately identify query-relevant chunks regardless of their position and the presence of noise, but in practice, it is often distracted by their placement and the unavoidable noise from the retrieval step. 

To bridge this gap, our training framework \algname{} adopts \emph{preference alignment} \cite{ethayarajh2024kto, guan2024deliberative, rafailov2024direct}, introducing explicit training signals to steer the generation model toward desired behaviors. The effectiveness of this approach hinges on the availability of \emph{high-quality} alignment data, as unreliable feedback can directly undermine the model’s robustness. To ensure quality, we construct a carefully curated pairwise comparison dataset in which the "chosen" response remains correct despite noisy or misplaced chunks in the input, while the "rejected" one fails. Learning from these contrastive examples enables the model to reliably resist retrieval-induced noise and positional variance, as illustrated in Figure~\ref{fig:overview}.


In detail, we simulate \emph{noisy} inputs that closely mirror real-world RAG generation scenarios, constructing a large-scale training dataset that reflects realistic retrieval conditions in which results contain both relevant and irrelevant information. Firstly, we generate question and answer pairs using LLMs from multi-domain source datasets, including HotPotQA (wiki), MS-MARCO (web search), PubMed (medical), CNNDM (news), and GovReport (report). Secondly, for each query,  we collect "relevant chunks" that contain the correct answer, along with multiple "irrelevant chunks" filtered from the chunk set obtained by a retrieval strategy.
To simulate realistic noisy input, chunks are mixed and shuffled with up to 25 sampled irrelevant chunks and a relevant chunk. This design mirrors critical challenges in the generation step of RAG, namely uncertain placement, as well as information noisiness.

Based on these constructed inputs, we generate corresponding outputs and collect feedback to create pairwise data for preference alignment. Here, constructing \emph{high-quality} feedback, \emph{i.e.}, chosen-rejected output pairs, is essential, as it provides explicit supervision that guides the model toward robust generation. By reinforcing preferred (chosen) outputs over dispreferred (rejected) ones given the noisy input, the model reduces forgetting and minimizes distraction from noisy retrieval results, ensuring more reliable answers in the generation stage. To enable this, we first collect candidate output completions using eight popular LLMs\footnote{Eight LLMs, varying in performance levels, are selected to ensure diverse response quality, enabling the construction of varied pairwise feedback for alignment tuning.}, by providing our noisy inputs for the question and answering (QA) task. Each model is prompted to generate outputs in the extract-then-generate style, where relevant evidence is explicitly extracted before composing the final answer. We then construct pairwise feedback by evaluating these candidate outputs using four widely adopted QA metrics in RAG settings \cite{fan2024survey, gao2023retrieval}: two inclusion-based metrics, namely \texttt{Accuracy (Acc)}, LLM-based evaluation (\texttt{LLMEval}); and two similarity-based metrics, namely \texttt{ROUGE-L}, and \texttt{BERTScore}. Outputs with high scores across these QA metrics are selected as "{chosen}," as they are considered robust to noisy input, whereas those with low scores are labeled as "{rejected}," since they represent undesired or incorrect answers (see Section \ref{sec:feedback_collection}).

Therefore, this feedback serves as direct supervision for preference alignment, such as DPO \cite{rafailov2024direct} and KTO \cite{ethayarajh2024kto}, guiding the model to produce high-quality extract-then-generate outputs even when the input includes irrelevant chunks or mispositioned relevant ones. That is, the model trained with this feedback directly couples evidence selection with answer generation, achieving greater robustness than approaches that rely on a separate, pre-generation compression module.
Our main contributions are: 

\begin{itemize}[leftmargin=*]
\item We introduce \algname{}, an extract-then-generate framework that trains LLMs to be resilient to information forgetting and noisiness, enabling robust generation in RAG systems.
\smallskip
\item We construct the first preference alignment training dataset, enabling models to learn to prioritize relevant information while effectively filtering out noise.
\smallskip
\item Our systematically constructed preference alignment dataset enables the model to achieve better Pareto-optimal performance, showing that balanced improvement over four QA metrics.
\smallskip
\item \algname{} outperforms independent filtering methods like \texttt{Recomp} and \texttt{CompAct}, demonstrating the benefit of dynamic, in-generation evidence selection over static, pre-filtered inputs.
\smallskip
\item \algname{} complements retrieval by addressing its limitations, improving the interpretation and use of retrieved content when deployed in real-world RAG pipelines.
\end{itemize}

\section{Related Work}
\label{sec:preliminary}

\paragraph{Retrieval in RAG} The retrieval is an essential process to fetch the most relevant text chunks to ground the responses to the given query. Two traditional approaches are employed for retrieval: \emph{sparse} retrieval, which relies on lexical-based methods such as BM25 \cite{robertson2009probabilistic}, and \emph{dense} retrieval, which uses text embeddings from both queries and text chunks \cite{zhao2024dense}. 
With the recent advance in RAG, significant efforts have been made to maximize retrieval performance. These include techniques: \emph{query rewriting} enriches the original query with semantically related terms to improve recall using LLMs \cite{gao2023precise, rashid2024progressive, wang2023query2doc, zhang2024mugi}; \emph{re-ranking} refines the initial retrieval results using more sophisticated models, often leveraging cross-encoders for better relevance estimation \cite{hwang2024dslr, reddy2024first, yurankrag}; and \emph{self-critique} iteratively verifies retrieved content for factual consistency \cite{asaiself, he2024retrieving, ye2024r} and can integrate web search for up-to-date information \cite{yan2024corrective}. 
Despite improvements in retrieval accuracy, hallucinations during generation caused by noisy retrieval results persist, highlighting the need for complementary research efforts \cite{islam2024open, laban2024summary}.

\paragraph{Generation in RAG} The generation is the crucial process of producing responses grounded in retrieved content. However, hallucinations still persist due to the inability of the LLM on noisy information \cite{cuconasu2024power}. In particular, Laban et al. \cite{laban2024summary} evaluated 50 RAG systems on the "Summary of a Haystack" benchmark, revealing that robust generation remains an {open challenge} even with high retrieval accuracy. To our knowledge, efforts to enhance the robustness of generation models against uncertain placement and information noisiness are limited. Instead, related efforts have primarily targeted the pre-generation stage, introducing an independent compression model before the generation model. 
ReComp \cite{xu2023recomp} compresses retrieved chunks into concise summaries to filter out noise and preserve only the most essential information using a fine-tuned contriver model. 
Similarly, CompAct \cite{yoon2024compact} and EXIT \cite{hwang2025exit} compress the retrieved chunks but rely on much larger LLMs. CompAct performs iterative abstractive compression with early termination, while EXIT adopts a parallel, context-aware extractive approach for sentence selection.
However, when compression is done as an independent step, the model sees only pre-filtered content and lacks feedback linking evidence selection to answer quality. This hinders alignment under noisy or uncertain retrieval. In contrast, integrating extraction into generation lets preference signals jointly shape evidence selection and answer formulation, enabling dynamic, query-aware focus during decoding.



\begin{table}[t]
\begin{center}
\footnotesize
\begin{tabular}{|L{8.0cm}|} \toprule
Ext2Gen Prompt \\ \midrule
You are an expert assistant trained to extract essential sentences from document chunks and generate answers based on the extracted sentences.
Your task is twofold:\\
- Extraction: Identify sentences that contribute to constructing a precise and accurate response to the given query.\\
- Generation: Formulate a concise and coherent answer based on the extracted sentences.

\vspace*{0.1cm}

\#\#\# Extraction Instruction:\\
- A query will be provided for you to answer.\\
- Extract only the sentences that contribute to forming an answer to the query.\\
- Ensure that the extracted sentences are sufficient to derive a correct and complete answer.\\
- If no relevant sentences are found in the provided chunks, return an empty list.

\vspace*{0.1cm}

\#\#\# Generation Instruction:\\
- Use the extracted sentences to generate a well-formed answer to the query. \\
- If no sentences are extracted, return "No Answer".

\vspace*{0.1cm}
\#\#\# Output Example:\\
Extracted Sentences:\\
- Sentence 1\\
- Sentence 2

\vspace*{0.1cm}

Answer: Your Answer

\vspace*{0.1cm}

\#\#\# Query: {\color{blue}\{query\}}

\vspace*{0.1cm}

\#\#\# Chunk List: {\color{blue}\{noisy chunk list\}}

\vspace*{0.1cm}

\#\#\# Output:
\\\bottomrule
\end{tabular}
\end{center}
\caption{{Base Prompt for Ext2Gen.} The prompt instructs the model to extract the essential sentences from document chunks and then to response to the query. }
\label{table:base_prompt}
\vspace*{-0.8cm}
\end{table}

%

\paragraph{Preference Alignment} 
The preference alignment (or optimization) process is essential for bridging the gap between human intent and the outputs generated by LLMs \cite{guan2024deliberative, wang2024comprehensive}. It serves as a critical mechanism to steer LLMs toward generating responses that align more closely with human expectations. Within this framework, preference optimization plays a central role by guiding models to prioritize responses preferred by human annotators over less desirable alternatives. Several optimization techniques for alignment have been proposed to achieve this, including PPO \cite{schulman2017proximal}, DPO \cite{rafailov2024direct}, and KTO \cite{ethayarajh2024kto}, as well as recent advances such as SimPO \cite{meng2024simpo}. These methods have been shown to be effective in aligning LLM behavior with human values, significantly reducing hallucinations, harmful outputs, and biased content \cite{wang2024comprehensive}.

\section{Alignment with Ext2Gen}
\label{sec:method}

\begin{table}[t]
\begin{center}
\footnotesize
\begin{tabular}{|L{8.0cm}|} \toprule
Data Generation: QA Generation Prompt \\ \midrule
You are a Question and Answer generation system. \\
Your task is to create a relevant query and provide a corresponding answer based on the given document chunk. \\
The query should be concise, clear, and directly relevant to the content of the document chunk. \\
The answer must be concise, factually grounded by the chunk, and formatted as either a phrase or a single sentence, aligned with one of the following categories:

\vspace*{0.1cm}

1. Fact-based: Generate a query that asks for specific details like dates, names, locations, etc., and provide a concise factual answer.\\
2. Instruction-based: Generate a query asking how to perform an action, and provide a concise step-by-step guide or instruction.\\
3. Definition or Explanation: Generate a query asking for a brief definition or explanation of a term or concept, and provide a clear explanation.\\
4. Opinion: Formulate a query that seeks advice or a recommendation based on the document content, and provide a brief opinion or recommendation.\\
5. Yes/No: Create a yes/no question based on the document chunk and answer it with "Yes" or "No."

\vspace*{0.1cm}

Your output must include a single query and its corresponding answer in JSON format:\\
\{\\
  "query": "your query belong to the five categories",\\
  "answer": "your answer"\\
\}

\vspace*{0.1cm}

\#\#\# Document Chunk: {\color{blue}\{target chunk\}}

\vspace*{0.1cm}

\#\#\# JSON Output:
\\\bottomrule
\end{tabular}
\end{center}
\caption{\textbf{QA Generation Prompt.} The prompt instructs the model to generate one of five QA types, covering query styles ranging from short-form to long-form QA formats. }
\label{table:qa_prompt}
\vspace*{-0.8cm}
\end{table}

\subsection{Overview}
To achieve the desired robustness in RAG generation models, direct model training is essential, as  prompt engineering with LLMs proves insufficient even with the sophisticated prompt \cite{liu2024enhancing, song2024learning}.
%
To this end, we explicitly teach LLMs to extract key sentences from a given set of chunks, encouraging a evidence reasoning process where the model first identifies grounding sentences for the query before generating the final answer.
At a high level, our framework consists of three main stages, as illustrated in Figure \ref{fig:overview}. First, we construct training data by generating candidate outputs for noisy inputs and collecting corresponding feedback through automated evaluation. Next, we form high-quality preference pairs (chosen vs. rejected) based on this feedback. Finally, we train the model using these pairs via preference optimization to align generation behavior with robust, evidence-grounded outputs.

\paragraph{Step 1: Data Generation.} We begin with a collection of {<Question, Answer>} pairs sourced from multiple domains, where each QA pair is aligned with a set of mixed text chunks containing both relevant and irrelevant context. Then, we filter the collected data based on two perspectives: "answer validity," ensuring the answer in QA is derivable from the relevant chunk; and  "chunk validity," confirming that none of the noisy chunks can infer the answer.

\paragraph{Step 2: Feedback Collection.} We collect a diverse set of possible output completions, achieved by prompting eight LLMs with the \algname{} prompt (in Table \ref{table:base_prompt}) along with the filtered query and mixed chunks. The outputs are then validated for format compliance, ensuring they include both extractive sentences and the final answer. To assign ratings of each output completion, we apply four popular QA metrics. Based on these ratings, we construct a set of pairwise feedback for each query by comparing multiple completions.

\paragraph{Step 3: Preference Optimization.} We train LLMs using  preference optimization, leveraging pairwise feedback to minimize the alignment gap in generation for RAG. We investigate the effectiveness of several optimization techniques, including supervised fine-tuning (SFT), DPO, KTO, and SimPO.


\subsection{Dataset Generation}

\begin{table}[t]
\begin{center}
\footnotesize
\begin{tabular}{|L{8.0cm}|} \toprule
Filtering: Answer--Chunk Validity Check Prompt \\ \midrule
You are responsible for evaluating whether the provided answer to the query can be derived from the given chunk.

\vspace*{0.1cm}

\#\#\# Instructions:\\
1. Analyze the provided answer in response to the query, using the information available in the chunk.\\
2. If the answer can be fully derived from the chunk, respond with "Supported".\\
3. If the answer cannot be fully derived from the chunk, respond with "Not Supported".

\vspace*{0.1cm}

Your output must be in JSON format. The output should be a dictionary whose a single key is "response".\\
\{\\
  "response": "Supported",\\
\}

\vspace*{0.1cm}

\#\#\# Query: {\color{blue}\{query\}}

\vspace*{0.1cm}

\#\#\# Answer: {\color{blue}\{answer\}}

\vspace*{0.1cm}

\#\#\# Chunk: {\color{blue}\{target chunk\}}

\vspace*{0.1cm}

\#\#\# JSON Output:
\\\bottomrule
\end{tabular}
\end{center}
\caption{\textbf{Answer--Chunk Validity Check Prompt.} For "answer validity ," if the answer is not supported by the relevant chunk, the corresponding QA pair is removed. For "noisy chunk validity," if the answer is supported by a noisy chunk, that noisy chunk is removed from the noisy chunk set.}
\label{table:validity_check}
\vspace*{-0.8cm}
\end{table}

\paragraph{QA Generation.} 
The diversity of source domains is essential for building comprehensive QA datasets, as it helps the model generalize across a wide range of contexts. For domains with existing high-quality human-generated QA pairs, HotPotQA (Wikipedia) \cite{yang2018hotpotqa} and MS-MARCO (web search) \cite{craswell2021ms}, we directly sample 4K QA pairs from each dataset. In the case of MS-MARCO, we further categorize the queries based on the original dataset's definitions into two types: long-form (“description”) and short-form. We sample 2K QA pairs for each type, leading to a total of 8K QA pairs from HotPotQA and MS-MARCO combined.

For the remaining domains, PubMed (Medical) \cite{cohan2018discourse}, CNNDM (News) \cite{nallapati2016abstractive}, and GovReport (Report) \cite{huang2021efficient}, where such QA annotations are not available, we generate 4K {<Question, Answer>} pairs per domain using GPT-4o. This results in a balanced and diverse set of QAs across five distinct domains. The prompt used for generating queries in this case is shown in Table \ref{table:qa_prompt}.

\paragraph{Chunk Collection.} To simulate the input used during the generation stage of RAG, we construct input contexts by retrieving document chunks for each query in the QA pairs. Specifically, for each query, we retrieve a set of document chunks that collectively serve as the generation model’s input during answer generation. The chunk originally used to generate the QA pair is designated as the "relevant" chunk. To obtain "irrelevant" chunks, we first store all available text chunks in a vector database using ChromaDB and perform dense retrieval with the multilingual-e5-large-instruct model \cite{wang2024multilingual}, retrieving the top-50 chunks for each query.

After removing any chunks identical to the relevant one, the remaining results are treated as "noisy" chunks. Here, we refer to them as noisy because some retrieved chunks that are not removed by exact matching may still contain relevant information.
This process yields 4K QA–chunk instances per dataset, each containing one QA pair and its corresponding chunk set, totaling 20K instances across five source datasets.

%


\paragraph{Data Filtering.} 

Our dataset includes both human- and LLM-generated QA pairs. However, for LLM-generated QA pairs, there is a risk of hallucination, which can introduce undesirable biases into the dataset \cite{das2024under, li2023synthetic}. To mitigate this issue, we carefully inspect the QA pairs and their associated chunks, since hallucinated answers may not be supported by the relevant chunks. Furthermore, beyond hallucinations, some chunks labeled as "noisy" may still contain information that supports the correct answer, despite not being explicitly marked as relevant.
Therefore, to mitigate their adverse effects on alignment training, we introduce an additional filtering step to refine the initial QA pairs and their associated noisy chunk set. In this step, we use Llama3.3-70b-instruct as the filtering model to reduce potential self-bias from GPT-4o, as using the same LLM for both generation and evaluation can lead to biased judgments favoring its own outputs \cite{xu2024pride}.

\begin{table}[t!]
\footnotesize
\begin{center}
\setlength{\tabcolsep}{18pt}
\renewcommand{\arraystretch}{1.1}
\begin{tabular}{|l|l|}
\toprule
\multicolumn{1}{|l|}{Model   Name}        &  Checkpoints   \\ \midrule                               
Llama3.2-3b-instruct    
& meta-llama/Llama-3.2-3B-Instruct    \\
Llama3.1-8b-instruct              
& meta-llama/Llama-3.1-8B-Instruct   \\
Nemo-12b-instruct               
& nvidia/Mistral-NeMo-12B-Instruct  \\
Gemma2-27b-instruct              
& google/gemma-2-27b-it   \\
Wizardlm2-8x22b             
&  alpindale/WizardLM-2-8x22B  \\
GPT-4o-mini                          
& gpt-4o-mini-2024-07-18 (OpenAI)           \\
Llama3.3-70b-instruct         
& meta-llama/Llama-3.3-70B-Instruct    \\
Llama3.1-405b-instruct             
&  meta-llama/Llama-3.1-405B-Instruct   \\
\bottomrule
\end{tabular}
\caption{Checkpoints: we use Hugging Face checkpoints for open-source models and OpenAI’s paid APIs for GPT-4o.}
\label{table:llm-source}
\end{center}
\vspace*{-0.8cm}
\end{table}

\begin{table*}[t]
\begin{center}
\small
\begin{tabular}{|L{1.1cm} |X{1.53cm} |X{1.53cm} |X{1.43cm} |X{1.63cm} |X{2.03cm} |X{1.63cm} |X{1.73cm} |X{1.8cm} |}\toprule
LLMs & \!\!Llama3.2-3b*\!\! & \!\!Llama3.1-8b*\!\! & Nemo-12b* & \!\!\!\!Gemma2-27b*\!\!\!\! & \!\!\!\!\!\!Wizardlm-2-8x22b\!\!\!\!\!\! & \!\!GPT-4o-mini\!\! & Llama3.3-70b* & \!\!\!\!Llama3.1-405b*\!\!\!\! \\ \midrule 
Chosen\!\! & ~~7.9\% & ~~9.6\% & 11.0\% & 11.0\% & 12.7\% & 14.4\% & 16.5\% & 16.9\% \\ 
Rejected\!\! & 22.7\% & 17.5\% & 15.5\%  & 15.1\% & 11.9\%  & ~~6.7\% & ~~5.0\% & ~~5.6\% \\ \bottomrule
\end{tabular}
\end{center}
\vspace*{0.0cm}
\caption{Distribution of "chosen" and "rejected" output completions for eight LLMs (in Table \ref{table:llm-source}), with {Rule 2} applied for pairwise comparison. The models are sorted in ascending order from left to right based on MMLU \cite{hendrycksmeasuring} and OpenLLM \cite{myrzakhan2024open} benchmark scores. That is, stronger LLMs positioned further to the right. $^{*}$ denotes the instruct-tuned version.} 
\label{table:distribution_output}
\vspace*{-0.6cm}
\end{table*}

Specifically, we prompt the Llama3.3 model with the validity check prompt in Table \ref{table:validity_check} to assess: \emph{"Answer Validation,"} where the answer is evaluated to ensure that it is fully derived from the relevant chunk—if not, the QA pair is filtered out as incorrectly generated (\emph{i.e.}, hallucination); and \emph{"Chunk Validation,"} where each chunk in the noisy set is checked to confirm that the answer cannot be derived from it—if it can, the chunk is removed from the noisy set (\emph{i.e.}, incorrect labels). 
We use the same prompt for both checks, as they perform the same task of verifying whether a given chunk can support the answer, regardless of its label.
This process yields 18K QA pairs with clearly labeled chunks as either "relevant" or "irrelevant," with the answers serving as "true" references.

\paragraph{Input Consolidation.} For the 18K subset, each query and its irrelevant chunk set are used to construct the input that mimics the generation step in RAG.
To better reflect noisy retrieval scenarios, we combine the relevant chunk with up to 25 chunks uniformly sampled from the irrelevant set, then randomly shuffle the combined chunks to form a chunk list. This list simulates the variability of Top-$k$ retrieval in RAG, capturing both the uncertain placement of the relevant chunk and the presence of varying amounts of irrelevant information.
The final input prompt for answer generation is constructed using the query and the shuffled chunk list, following the \algname{} prompt format shown in Table \ref{table:base_prompt}.

In the next step, we use the constructed noisy input to generate both chosen (desirable) and rejected (undesirable) outputs, guiding alignment tuning so the model learns to produce desirable responses even under noisy retrieval conditions.

\subsection{Feedback Collection}
\label{sec:feedback_collection}

\paragraph{Output Generation.} With the simulated inputs, we collect multiple output completions using the \algname{} prompt in Table \ref{table:base_prompt} by prompting eight LLMs with varying performance levels. Refer to the full list of LLMs in Table \ref{table:llm-source}. 
Among these candidate completions, higher-quality responses are labeled as "chosen" while lower-quality or flawed ones are labeled as "rejected." 
This model diversity is critical for effective alignment, as it enables us to gather responses of varying quality for the same input, which is essential for constructing informative pairwise feedback \cite{chaudhari2024rlhf, song2024preference, song2024learning}.
As a result, these models generate a total of 144K input–response pairs (18K noisy input instances $\times$ 8 models) with varied quality levels. 

\paragraph{Output Compliance.} To align with the expected \algname{} output, we normalize LLMs' output completions by removing unintended ones, such as those containing only extracted sentences, direct answers to the query without sentences extracted, or outputs that follow an incorrect format. In alignment, this process helps the model maintain the consistent completion format as:

\begin{tcolorbox}[colback=gray!10, colframe=black!80, title=Ext2Gen Output Completion]
\vspace*{-0.2cm}
Extracted Sentences:\\
- {sentence 1}\\
- {sentence 2}\\

\vspace*{-0.3cm}
Answer: {generated answer}
\vspace*{-0.2cm}
\end{tcolorbox}

\paragraph{Feedback Composition.} 
We configure pairwise feedback for preference alignment by contrasting the correctness of multiple output completions. We use four metrics to evaluate the correctness of the output from two perspectives: Accuracy \texttt{(Acc)} and \texttt{LLMEval} for assessing the {"inclusion"} of the true answer in the generated one, and \texttt{ROUGE-L} and \texttt{BERTScore} for measuring {lexical} and semantic {"similarity"} between the true and generated one.

\begin{itemize}[leftmargin=*]
\item \textbf{Accuracy (Acc)} checks whether the true answer is included in the predicted response. Unlike exact match, it allows for partial inclusion, making it suitable for RAG evaluation.
\item \textbf{LLMEval} is Similar to \texttt{Acc}, but uses GPT-4o to assess answer correctness beyond lexical overlap, enabling more context-aware evaluation (see the prompt in Table~\ref{table:llm_eval}).    
\smallskip\smallskip
\item \textbf{ROUGE} \cite{lin2004rouge} measures lexical overlap via the F1-score of \texttt{ROUGE-L}, focusing on the longest common subsequence to capture both key phrase and word order similarity.   
\smallskip\smallskip
\item \textbf{BERTScore} \cite{zhangbertscore} computes semantic similarity using contextual embeddings, measuring token-level cosine similarity. 
\end{itemize}

\smallskip\smallskip
Based on the rated scores, we define two rules for selecting "chosen" and "rejected" output completions. The first rule considers only inclusion-based metrics, which are known to be more critical than similarity metrics in RAG settings \cite{yu2024evaluation}. The second rule incorporates all metrics, prioritizing them in the following order:

\smallskip\smallskip
\noindent {\textbf{{Rule 1: Inclusion-only.}}} 
This considers only binary inclusion metrics (\texttt{Acc} and \texttt{LLMEval}), where 1 indicates the generated answer includes the true answer, and 0 indicates it does not.
An output is considered {"chosen"} if either metric equals 1. The condition holds for any chosen output $i$:
\begin{equation}
{\rm \text{Acc}}_i + {\rm \text{LLMEval}}_i \geq 1,
\label{eq:chosen}
\end{equation}
indicating at least one of the inclusion metrics confirms the presence of the true answer, where $\{\rm \sc metric\}_i$ denotes the metric score for the output $i$. 
Then, for any chosen output $i$, another output $j$ is considered a {"rejected"} output if:
\begin{equation}
{\rm \text{Acc}}_i + {\rm \text{LLMEval}}_i > {\rm \text{Acc}}_j + {\rm \sc \text{LLMEval}}_j,
\label{eq:rule1_reject}
\end{equation}
ensuring that the chosen one has a stronger inclusion signal than the rejected one.

\begin{table}[t]
\begin{center}
\footnotesize
\begin{tabular}{|L{8cm}|} \toprule
LLMEval Prompt \\ \midrule
Your task is to evaluate the correctness of the predicted answer based on the true answer. 

\vspace*{0.1cm}

\#\#\# Instructions:\\
- Read the QUERY and then compare the ANSWER and the Predicted ANSWER.\\
- Check if the Predicted Answer includes the core content of the True Answer (True/False in text).

\vspace*{0.1cm}

\#\#\# QUERY: {\color{blue} \{query\}}

\vspace*{0.1cm}

\#\#\# TRUE ANSWER: {\color{blue} \{true answer\}}

\vspace*{0.1cm}

\#\#\# Predicted ANSWER: {\color{blue} \{predicted answer\}}

\vspace*{0.1cm}

\#\#\# Output Format:
\{
    "Correctness": "True or False"
\}

\vspace*{0.1cm}

\#\#\# Output (Only JSON):
\\\bottomrule
\end{tabular}
\end{center}
\caption{\textbf{LLMEval Prompt.} This prompt is used to verify the faithfulness of the generated answer and the correctness of relevant and irrelevant chunks.}
\label{table:llm_eval}
\vspace*{-0.8cm}
\end{table}

\begin{table*}[t]
\begin{center}
\begin{tabular}{|X{2.0cm} |X{1.18cm} X{1.18cm} X{1.18cm} X{1.18cm} |X{1.18cm} || X{1.18cm} X{1.18cm} X{1.18cm} X{1.18cm} |X{1.18cm}|} \toprule
Backbone & \multicolumn{5}{c||}{Llama3.1-8b-instruct} & \multicolumn{5}{c|}{Llama3.2-3b-instruct} \\ \midrule
\!Metric & Acc & \!\!\!LLMEval\!\!\! & \!\!\!ROUGE\!\!\! & BERT & Avg. & Acc & \!\!\!LLMEval\!\!\! & \!\!\!ROUGE\!\!\! & BERT & Avg. \\ \midrule
\!\texttt{Ideal} & 0.439 & 0.918 & 0.339 & 0.881 & 0.644 & 0.446 & 0.877 & 0.310 & 0.876 & 0.627\\ \midrule
\!\texttt{Default} & 0.341 & 0.733 & 0.212 & 0.859 & 0.536 & 0.286 & 0.595 & 0.162 & 0.849 & 0.473 \\ 
\!\texttt{SFT-Best} & 0.363 & 0.763 & 0.282 & 0.871 & 0.570 & 0.295 & 0.649 & 0.226 & 0.861 & 0.508\\ 
\!\textbf{Ext2Gen-R1}\! & \textbf{0.481} & \textbf{0.889} & 0.212 & 0.860 & 0.610 & \textbf{0.401} & \textbf{0.773} & 0.179 & 0.854 & 0.552 \\
\!\textbf{Ext2Gen-R2}\! & 0.463 & 0.860 & \textbf{0.370} & \textbf{0.885} & \textbf{0.644} & 0.390 & 0.750 & \textbf{0.228} & \textbf{0.860} & \textbf{0.557} \\\bottomrule
\end{tabular}
\end{center}
\caption{Evaluation results of five methods using Llama3.1-8b-instruct (left) and Llama3.2-3b-instruct (right) as the backbone, where ROUGE and BERT refer to ROUGE-L and BERTScore, respectively. } 
\label{table:exp1-main}
\vspace*{-0.7cm}
\end{table*}

\smallskip\smallskip
\noindent {\textbf{{Rule 2: Inclusion $\rightarrow$ Similarity.}}} This rule considers both inclusion and similarity metrics, giving higher priority to \texttt{Acc} and \texttt{LLMEval} over \texttt{ROUGE-L} and \texttt{BERTScore}. The basic criteria for determining chosen and rejected outputs are the same as in Rule 1, defined in Eqs.~(\ref{eq:chosen}) and (\ref{eq:rule1_reject}).
However, for rejected outputs, in addition to the condition in Eq.~(\ref{eq:rule1_reject}), we introduce an additional criterion to generate more chosen-rejected feedback pairs when two outputs, \(i\) and \(j\), have identical inclusion scores, \emph{i.e.}, ${\rm \sc Acc}_i = {\rm \sc Acc}_j~{\rm and}~{\rm \sc LLMEval}_i = {\rm \sc LLMEval}_j$. Specifically, even if output \(j\) has the same inclusion score as the chosen output \(i\), it is considered {"rejected"} if: 
\begin{equation}
{\rm \text{ROUGE-L}}_i + {\rm \text{BERTScore}}_i > {\rm \text{ROUGE-L}}_j +  {\rm  \text{BERTScore}}_j + \epsilon.\!\!\!
\label{eq:rule2_reject}
\end{equation}
This guarantees that outputs are preferred not only for including the true answer but also for exhibiting higher lexical and semantic similarity to it. The $\epsilon$ is set to 0.30 for the chosen one to have a sufficiently higher similarity score than the rejected one.
By applying the two rules, we construct 120K feedback pairs under Rule 1, which emphasizes inclusion-based metrics, and 150K pairs under Rule 2, which incorporates both inclusion and similarity metrics for a more holistic evaluation.

Table \ref{table:distribution_output} shows the proportion of each LLM's outputs judged as "chosen" or "rejected" in the 150K feedback set (by Rule 2). While stronger LLMs (on the right) are more frequently chosen, not all of their outputs are preferred, and weaker LLMs (on the left) often produce better completions. This supports our strategy of sampling candidates from a diverse pool of LLMs, as it enables more nuanced pairwise selection based on the real correctness of responses rather than the identity of the model. In doing so, {our feedback composition naturally incorporates diverse LLM outputs into alignment training.} The distribution under Rule 1 follows a consistent trend, as the 150K pairs form a superset that includes all 120K pairs.

\subsection{Preference Optimization}  

We use pairwise feedback, where each input—consisting of a query and its associated chunk list—is paired with a chosen output that is preferred and a rejected output that is less preferred.
We train the generation model directly to favor the chosen output over the rejected one. Although GPT-4o was partly used to generate QA pairs during data construction, none of its responses are included in the training dataset. Instead, training responses come from a diverse set of models listed in Table \ref{table:llm-source}, ensuring the model avoids imitating any single LLM and maintains stylistic diversity.

For alignment tuning, we primarily use Llama3.2-3b-instruct and Llama3.1-8b-instruct as generation backbones.
In our experiments, we explore seven training setups, primarily based on SFT and DPO, to evaluate how effectively our constructed pairwise feedback improves the robustness of the underlying backbones:

\smallskip
\noindent
$\bullet$ {\bf SFT-Best}: We first identify the best output for each query from the eight LLMs, selecting the one with the highest average score across four QA metrics. This output is then used as the unique reference completion for SFT.

\smallskip
\noindent
$\bullet$ {\bf SFT-\{Metric\}}: Similar to \texttt{SFT-Best}, but the best output is selected based on a single metric rather than the average of all four. This setup includes four more SFT variants: \texttt{SFT-Acc}, \texttt{SFT-LLMEval}, \texttt{SFT-ROUGE}, and \texttt{SFT-BERT}.

\smallskip
\noindent
$\bullet$ {\bf Ext2Gen-\{Rule\}}: Unlike the SFT variants, which rely on a single reference, we leverage \emph{multiple} pairwise feedback instances as optimization signals, even for the same query. We optimize our model using DPO based on the two feedback composition rules separately, resulting in two models: \texttt{Ext2Gen-R1} uses feedback from inclusion-based metrics only (Rule 1), while \texttt{Ext2Gen-R2} incorporates both inclusion and similarity metrics (Rule 2).
%
In addition to DPO, other alignment tuning methods can be applied. We compare DPO with KTO \cite{ethayarajh2024kto} and SimPO \cite{meng2024simpo} in Section \ref{sec:exp_sft_variants}.





\section{Evaluation}
\label{sec:exp}

This section presents two evaluations: (i) Robustness improvement of LLM backbones to chunk misplacement and information noisiness after alignment tuning with \algname{} (see Section \ref{sec:robustness_eval}), including the comparison with existing methods that extract relevant sentences prior to RAG generation; and (ii) Deployment of \algname{} models in a real RAG environment (see Section~\ref{sec:deployment_eval}).

\begin{figure*}[t!]
\begin{center}
\includegraphics[width=15.8cm]{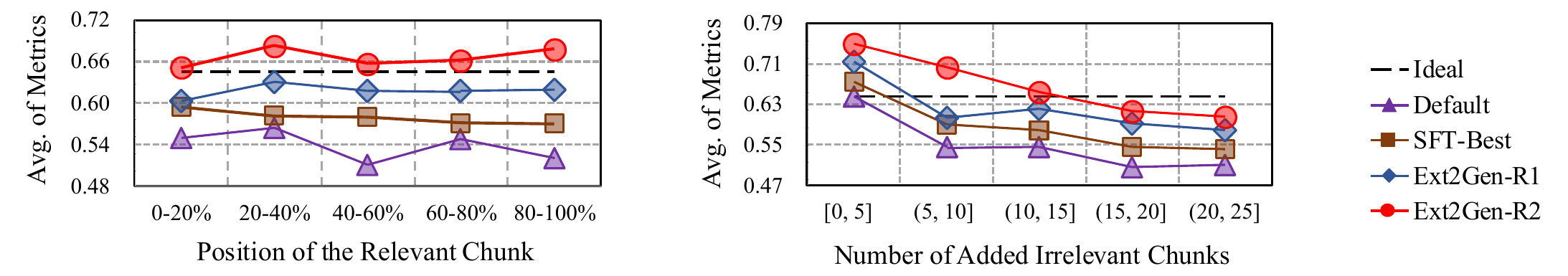}
\end{center}
\vspace*{-0.4cm}
\caption{Robustness to \textbf{(left)} relevant chunk position (moving down as it shifts right) and \textbf{(right)} the number of added irrelevant chunks (increasing noise level to the right). Results are based on the Llama3.1-8b-instruct backbone. }
\label{fig:exp-robustness-8b}
\vspace*{-0.3cm}
\end{figure*}

\subsection{Robustness Evaluation}
\label{sec:robustness_eval}

\paragraph{Configuration.} Since our goal is to directly enhance the robustness of the generation backbone, we primarily compare our two main models, namely \texttt{Ext2Gen-R1/R2}, with other variants trained with or without SFT. For both SFT and DPO, we mainly fine-tune Llama3.2-3b-instruct and Llama3.1-8b-instruct using QLoRA \cite{dettmers2024qlora} on four NVIDIA H100 GPUs. For consistency across all setups, the training process spans 9,000 steps, employing AdamW as the optimizer with a batch size of 32, an initial learning rate of 5e-6, and a weight decay of 0.05. For evaluation metrics, we employ the four metrics, \texttt{Acc}, \texttt{LLMEval}~(using GPT-4o), \texttt{ROUGE-L}, and \texttt{BERTScore}. 

Additionally, we compare our models with approaches that rely on an independent compression module prior to generation, such as \texttt{Recomp} \cite{xu2023recomp}, \texttt{CompAct} \cite{yoon2024compact}, and \texttt{EXIT} \cite{hwang2025exit} in Section \ref{sec:exp_compression}.

\paragraph{Test Dataset.} 
We construct the test set using the same pipeline as the \algname{} training set to assess robustness in QA generation, but with the "test split" of the five source datasets. Since only the input is required, the process in Figure \ref{fig:overview} runs only up to the input consolidation step for the test split. This results in a total of 1K QA pairs, with 200 QA pairs sampled or generated from each of the five source datasets. Note that in the \algname{} input prompt, each query is paired with a list of chunks containing both relevant chunks and up to 25 irrelevant ones. 

\vspace*{-0.024cm}
\subsubsection{Main Results}
\label{sec:exp1_main}
\vspace*{-0.024cm}

Table \ref{table:exp1-main} summarizes the generation performance of five models for the test set. \texttt{Default} (base model) refers to the results obtained using the \algname{} prompt in Table \ref{table:base_prompt} without preference alignment, neither SFT nor DPO is applied,  while \texttt{Ideal} represents those obtained with \texttt{Default} when only relevant chunks are provided as the chunk list.

Firstly, {the base model, Default, is highly sensitive to information forgetting and information noisiness}, experiencing significant performance drops; particularly in smaller models such as Llama3.2-3b-instruct. This vulnerability is especially critical in scenarios that prioritize compact models, such as RAG.    

Secondly, {alignment with Ext2Gen significantly boosts generation scores across all QA metrics}. Notably, leveraging constructed pairwise feedback leads to substantially greater improvements, as demonstrated by the \algname{} series outperforming \texttt{SFT-Best}, which relies solely on single best outputs without pairwise comparisons. See Section \ref{sec:exp_robustness} for detailed analysis.

\begin{table}[t]
\begin{center}
\footnotesize
\begin{tabular}{|L{8cm}|} \toprule
Filtering Prompt \\ \midrule
Extract key sentences from the retrieved documents to create an extractive summary that can be used to answer the question.

\vspace*{0.1cm}

\#\#\# Question:  {\color{blue} \{query\}}

\vspace*{0.1cm}

\#\#\# Retrieved documents: {\color{blue} \{noisy chunk list\}}

\vspace*{0.1cm}

\#\#\# Extractive summary:
\\\bottomrule
\end{tabular}
\end{center}
\caption{{Filtering Prompt.} This prompt extracts key sentences from noisy chunks using GPT-4o before generation.}
\label{table:filtering_gpt}
\vspace*{-0.8cm}
\end{table}

Lastly, \texttt{Ext2Gen-R2} reveals that {balancing inclusion and similarity metrics in feedback composition leads to better Pareto alignment}, resulting in the best model based on the average score.\footnote{Averaging multiple metrics is a common way to select the best model in multi-objective optimization, where balanced improvements reflect Pareto optimality \cite{yang2024rewards}.} With Llama3.1-8b-instruct, \texttt{Ext2Gen-R2} demonstrates strong robustness, achieving performance nearly indistinguishable from the noise-free \texttt{Ideal} model, even when up to 25 irrelevant chunks are added to the input. Meanwhile, \texttt{Ext2Gen-R1} surpasses \texttt{Ext2Gen-R2} on inclusion metrics but lags behind in similarity metrics.

\subsubsection{Robustness against Relevant Chunk Position and Information Noisiness}
\label{sec:exp_robustness}

Figure \ref{fig:exp-robustness-8b} provides a detailed visualization of the results summarized in Table \ref{table:exp1-main}. It shows how the average score (Avg.) across four metrics changes depending on the position of the relevant chunk (left) and the number of irrelevant chunks added (right) within the input prompt. Notably, \texttt{Ideal} maintains a constant score unaffected by either chunk misplacement or information noisiness, since its input contains only relevant chunks.

\begin{table}[t]
\begin{center}
\small
\begin{tabular}{ |L{1.4cm} | X{1.3cm} | X{0.9cm} X{0.90cm} X{0.9cm} X{0.9cm} | } \toprule
\!Metric & \!\!\!Filter Stage\!\!\! & Acc & \!\!\!LLMEval\!\!\! & \!\!\!ROUGE\!\!\! & BERT  \\ \midrule
\!\texttt{Default} & - & 0.341 & 0.733 & 0.212 & 0.859  \\ \midrule
\!\texttt{Recomp}  & Pre-gen. & 0.248 & 0.500 & 0.243 & 0.862  \\ 
\!\texttt{CompAct}  & Pre-gen. & 0.343 & 0.736 & 0.300 & 0.874  \\ 
\!\texttt{EXIT}  & Pre-gen. & 0.360 & 0.751 & 0.340 & 0.881  \\ 
\!\texttt{GPT-Filter}  & Pre-gen. & 0.399 & 0.840 & 0.345 & 0.881  \\ \midrule
\!\texttt{Ext2Gen-R2}\!\!\! & In-gen. & \textbf{0.463} & \textbf{0.860} & \textbf{0.370} & \textbf{0.885}  \\ \bottomrule
\end{tabular}
\end{center}
\caption{Comparison of \algname{} with other text compression methods applied before the generation stage in RAG, using Llama3.1-8b as the generation backbone for all methods. } 
\label{table:exp1-comp}
\vspace*{-0.7cm}
\end{table}

For relevant chunk position, \texttt{Ext2Gen-R2} consistently outperforms all other methods across every position of the relevant chunk within the input, even surpassing the \texttt{Ideal} baseline. This demonstrates its strong ability to adapt to shifts in the location of critical information, which is crucial in real-world retrieval scenarios where relevant content may appear unpredictably. Although \texttt{SFT-Best} and \texttt{Ext2Gen-R1} also show improvements over the \texttt{Default} model, their robustness to positional changes is less pronounced and more variable.
Similarly, when facing information noisiness caused by the addition of irrelevant chunks, \texttt{Ext2Gen-R2} exhibits superior resistance to performance degradation. It maintains significantly higher scores compared to all other baselines, reflecting its enhanced capability to filter noise and focus on relevant evidence even in challenging, noisy inputs. {This combination of adaptability to both chunk misplacement and input noisiness highlights the effectiveness of the \texttt{Ext2Gen-R2} alignment approach.}

\subsubsection{Comparison with Compression Method}
\label{sec:exp_compression}

We focus on directly enhancing the generation model's robustness itself. While differing in intent, \texttt{Recomp} \cite{xu2023recomp}, \texttt{CompAct} \cite{yoon2024compact}, and \texttt{EXIT} \cite{hwang2025exit} adopts independent filtering (compression) modules between retrieval and generation to extract only the most relevant information. Although not targeting the generation model directly, there pre-generation filtering strategies are conceptually related to our work, as it also seeks to reduce the influence of noise or irrelevant input on generation quality.
Thus, we compare \algname{} with the three compression methods \footnote{We use the extractive variant of \texttt{Recomp}, built on Contriever \cite{izacardunsupervised}. The number of extracted sentences is set to three, which closely matches the average number produced by \algname{}, as seen in Table \ref{table:output_eval}. For CompAct and EXIT, we directly use their publicly available Hugging Face models, namely CompAct-7B and EXIT-Gemma-7B.} and another strong counterpart, \texttt{GPT-Filter}, using GPT-4o as the compression model directly using a specialized prompt in Table \ref{table:filtering_gpt}. Note that these methods compress the retrieved chunks in the \emph{pre-generation} stage, whereas \algname{} integrates sentence selection into the \emph{in-generation} stage through alignment tuning; thus, \algname{} does not need any independent models, directly linking evidence selection to answer quality.

Table \ref{table:exp1-comp} compares the QA performance of \algname{} with four pre-generation compression methods, where only the summarized text is provided to the generation model. Notably, \texttt{Ext2Gen-R2} outperforms pre-generation compression methods across all metrics. \texttt{Recomp} underperforms Default, likely due to loss of essential context during its separate compression step. While \texttt{CompAct} and \texttt{EXIT} enhance generation performance through compression, they still underperform compared to \texttt{GPT-Filter}, which leverages a powerful model, GPT-4o, for sentence selection. However, its performance still falls short of \texttt{Ext2Gen-R2}, suggesting that integrating filtering into the generation process rather than applying it externally can yield better results. In-generation filtering enables the model to dynamically condition its sentence selection while generating answers, rather than relying on a static, pre-selected context.

\begin{table}[t]
\begin{center}
\small
\begin{tabular}{|L{1.4cm} |X{0.75cm} X{0.75cm} |X{0.75cm} |X{0.75cm} X{0.75cm} |X{0.75cm}| }\toprule
Backbone  & \multicolumn{3}{c|}{Llama3.1-8b-instruct} &  \multicolumn{3}{c|}{Llama3.2-3b-instruct}\\ \midrule
Method  & Prec. & Recall & F1 & Prec. & Recall & F1 \\  \midrule
\!\texttt{Default} & 0.43 & 0.76 & 0.45 & 0.30 & 0.68 & 0.42 \\ 
\!\texttt{SFT-Best} & 0.50 & 0.75 & 0.60 & 0.41 & 0.69 & 0.51 \\ 
\!\texttt{Ext2Gen-R1}\!\!\!\!  &  0.46& \textbf{0.91} & 0.61  & 0.36& \textbf{0.86} & 0.51\\ 
\!\texttt{Ext2Gen-R2}\!\!\!\!  &\textbf{0.62}  &{0.81} & \textbf{0.70}  & \textbf{0.42}  &0.82 & \textbf{0.56} \\ \bottomrule
\end{tabular}
\end{center}
\caption{Precision (Prec.), recall, and F1-score of the extracted sentences in output generated by four models.} 
\label{table:pre_rec}
\vspace*{-0.7cm}
\end{table}

\begin{table}[t]
\begin{center}
\small
\begin{tabular}{|L{1.4cm} |X{1.3cm} | X{1.3cm} |X{1.3cm} |X{1.3cm}| }\toprule
Method  & Sentence Number & Words in Sentences & Words in Answer  & Latency (sec/query)\\  \midrule
\!\texttt{Default}  & 4.81 & 115 & 46 & 6.66 \\ 
\!\texttt{Ext2Gen-R1}\!\!\!\!  & 5.10  & 127 & 59  & 7.52   \\ 
\!\texttt{Ext2Gen-R2}\!\!\!\!  & {3.26} & 77  & {43} & {5.34} \\ \bottomrule
\end{tabular}
\end{center}
\caption{Statistics of Ext2Gen outputs (averaged): number of extracted sentences along with their word counts, answer word counts, and query processing latency (seconds per query). The test inputs are lengthy, averaging 2,161 words, which contributes to the several-second inference time.} 
\label{table:output_eval}
\vspace*{-0.7cm}
\end{table}

\begin{figure*}[t!]
\begin{center}
\includegraphics[width=16cm]{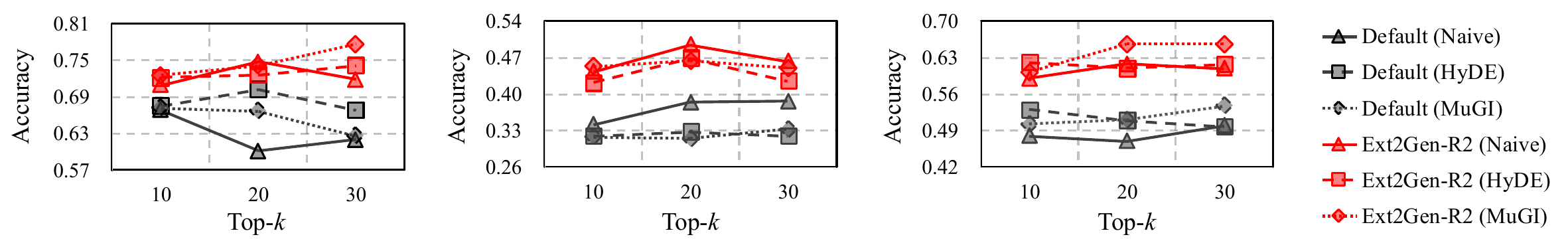}
\end{center}
\vspace*{-0.15cm}
{\small \hspace*{-2.45cm} (a) Natural Question (NQ). \hspace*{1.7cm}  (b) MS-MARCO. \hspace*{2.6cm} (c) HotPotQA.}
\vspace*{-0.25cm}
\caption{Accuracy of the Llama3.1-8b backbone fine-tuned with \algname{} in a RAG environment, evaluated across three retrieval approaches: naive dense retrieval (\texttt{Naive}) and its enhanced variants using query rewriting methods, \texttt{HyDE} \cite{gao2023precise} and \texttt{MuGI} \cite{zhang2024mugi}.}
\label{fig:deployment}
\vspace*{-0.15cm}
\end{figure*}

\subsubsection{Additional Analysis}
\label{sec:exp_sft_variants}

To further validate the effectiveness and versatility of \algname{}, we conduct a series of additional experiments. These include evaluating the quality and efficiency of extracted sentences, comparing different feedback tuning strategies, and testing the robustness of \algname{} across various optimization techniques and model backbones. Together, these analyses highlight the generalizability and practical advantages of \algname{}.

\paragraph{Quality of Extracted Sentences.}
The quality of extracted sentences is crucial for grounding the generated answer. In Table \ref{table:pre_rec}, we evaluate the quality of the extracted sentences using three metrics. \emph{Precision} measures the proportion of extracted sentences that originate from the relevant chunks, indicating how accurately the model selects content from the correct sources. \emph{Recall} measures the proportion of relevant chunks that contain at least one extracted sentence, reflecting how comprehensively the model covers the necessary information. Lastly, we report the \emph{F1-score}, the harmonic mean of precision and recall, to summarize overall extraction performance.

Since \texttt{Ext2Gen-R1} is trained with feedback composed solely based on inclusion-based metrics, it tends to generate longer responses to increase the chance of including the gold answer, resulting in high recall but low precision. In contrast, \texttt{Ext2Gen-R2} incorporates not only inclusion metrics but also lexical similarity metrics, encouraging responses that are not only accurate but also concise and similar to the reference. This results in a better balance between precision and recall, ultimately yielding the highest F1 score. Therefore, {the improved robustness of \texttt{Ext2Gen-R2} in Section \ref{sec:exp_robustness} is attributed to a well-balanced trade-off between precision and recall in sentence extraction}.


\paragraph{Output and Latency.} Table \ref{table:output_eval} presents the output statistics and latency of the \algname{} series compared to \texttt{Default}, measured on a single NVIDIA H100 with a batch size of 1. \texttt{Ext2Gen-R2}, which incorporates lexical similarity metrics such as ROUGE in addition to inclusion metrics, generates more concise extractions and answers. This not only improves output quality but also results in faster inference than \texttt{Default}. In contrast, \texttt{Ext2Gen-R1}, trained solely with inclusion-based feedback (\emph{e.g.}, \texttt{ACC} and \texttt{LLMEval}), produces more extracted sentences and longer outputs, leading to higher recall but increased latency and lower precision. 
These results suggest that considering both lexical similarity and inclusion metrics during feedback composition, as done in \texttt{Ext2Gen-R2}, is beneficial not only for improving robustness but also for reducing latency, thanks to more concise and focused extractions.

\begin{table}[t]
\begin{center}
\small
\begin{tabular}{|L{1.7cm} |X{1.22cm} X{1.22cm} X{1.22cm} X{1.22cm} |} \toprule
Method  & Acc & \!\!\!\!LLMEval\!\!\!\! & \!\!\!\!ROUGE\!\!\!\! & BERT \\\midrule
\!\texttt{SFT-Best} & 0.363 & 0.763 & 0.282 & 0.871  \\ \midrule
\!\texttt{SFT-Acc} & \textbf{0.376} & 0.763 & 0.282 & 0.871  \\ 
\!\texttt{SFT-LLMEval}\!\!\!\! & 0.360 & \textbf{0.777} & 0.220 & 0.861  \\ 
\!\texttt{SFT-ROUGE} & 0.368 & 0.748 & \textbf{0.284} & 0.869  \\ 
\!\texttt{SFT-BERT} & 0.357 & 0.744 & 0.280 & \textbf{0.873}  \\ \bottomrule
\end{tabular}
\end{center}
\caption{Comparison of SFT variants for alignment.} 
\label{table:sft-variants}
\vspace*{-0.8cm}
\end{table}

\begin{table}[t]
\begin{center}
\small
\begin{tabular}{|L{1.0cm} |X{1.4cm} X{1.4cm} X{1.4cm} X{1.4cm} |} \toprule
Size & Acc & \!\!\!\!LLMEval\!\!\!\! & \!\!\!\!ROUGE\!\!\!\! & BERT \!\!\\\midrule
\!\!\texttt{Default}\!\!\!\! & 0.34 & 0.73 & 0.21 & 0.86  \\\midrule
\!\!\texttt{DPO} & \!\!0.46~{\color{blue}(+0.12)}\!\! & \!\!\!\!0.86~{\color{blue}(+0.13)}\!\!\!\! & \!\!0.37~{\color{blue} (+0.16)}\!\! & \!\!0.86~{(+0.00)}\!\!  \\
\!\!\texttt{KTO} & \!\!0.44~{\color{blue}(+0.10)}\!\! & \!\!0.85~{\color{blue}(+0.12)}\!\!  & \!\!0.35~{\color{blue}(+0.14)}\!\!  & \!\!0.85~{\color{red}(-0.01)}\!\!   \\
\!\!\texttt{SimPO} & \!\!0.32~{\color{red}(-0.02)}\!\! & \!\!0.74~{\color{blue}(+0.02)}\!\! & \!\!0.34~{\color{blue}(+0.13)}\!\! & \!\!0.88~{\color{blue}(+0.02)}\!\!  \\\bottomrule
\end{tabular}
\end{center}
\caption{Comparison of \algname{} on the Llama3.1-8b-instruct backbone using \textbf{DPO}, \textbf{KTO}, and \textbf{SimPO}.} 
\label{table:comparision_kto_simpo}
\vspace*{-0.6cm}
\end{table}

\paragraph{SFT Variants.} Table \ref{table:sft-variants} compares \texttt{Ext2Gen-R2} with other SFT variants that use a single QA metric for feedback composition, where Llama3.1-8b-instruct is used. {Focusing on a single metric can introduce an alignment tax, leading to degraded performance on other QA metrics.} Each variant tends to excel in its targeted metric but underperforms on others. For example, \texttt{SFT-LLMEval} achieves the highest LLMEval score but the lowest ROUGE-L score, whereas \texttt{SFT-ROUGE} maximizes ROUGE-L at the expense of LLMEval.

\paragraph{Generalization to Optimization Method.}

With the rapid progress in preference optimization, several methods beyond DPO \cite{rafailov2024direct} have emerged. KTO \cite{ethayarajh2024kto} replaces paired preference data with binary feedback (“1” for desirable, “0” for not), so we convert our pairwise dataset into binary form for KTO. SimPO \cite{meng2024simpo}, in contrast, requires no reference model and uses the average log probability as an implicit reward, improving alignment with generation metrics while reducing computational and memory costs.
Table \ref{table:comparision_kto_simpo} compares the performance of \algname{} trained with three different optimization methods. All methods improve QA performance, with DPO achieving the highest gains in \texttt{Acc}, \texttt{LLMEval}, and \texttt{ROUGE-L}, which is why we adopt DPO as the primary method in our main experiments. KTO performs comparably to DPO, despite using only binary labels instead of paired preference data. This indicates that KTO may offer a more efficient alternative by simplifying feedback construction. In contrast, SimPO underperforms relative to both DPO and KTO.

\paragraph{Generalization to Other Backbone.}

We validate the generalization capability of our alignment pipeline by training it with Qwen2.5-3b-instruct as the backbone, instead of our two primary models, Llama3.2-3b-instruct and Llama3.1-8b-instruct. Table \ref{table:exp1-qwen} reports QA performance across four evaluation metrics for four approaches: \texttt{Ideal}, \texttt{Default}, \texttt{SFT-Best}, and \texttt{Ext2Gen-R2}; all trained on the Qwen backbone. Overall, \texttt{Ext2Gen-R2} consistently outperforms the baselines, with especially notable gains over \texttt{Default} in the two key metrics: \texttt{Acc} and \texttt{LLMEval}. These results confirm that our \algname{} framework is transferable across different architectures.

\begin{table}[t]
\begin{center}
\small
\begin{tabular}{ |L{1.5cm} |X{1.3cm} X{1.3cm} X{1.3cm} X{1.3cm} | } \toprule
\!Metric & Acc & \!\!\!LLMEval\!\!\! & \!\!\!ROUGE\!\!\! & BERT  \\ \midrule
\!\texttt{Ideal} & 0.417 & 0.877 & 0.271 & 0.871 \\ \midrule
\!\texttt{Default} & 0.258 & 0.516 & 0.141 & 0.843  \\ 
\!\texttt{SFT-Best} & 0.305 & 0.609 & 0.217 & 0.859 \\ 
\!\texttt{Ext2Gen-R2}\!\!\! & \textbf{0.318} & \textbf{0.649} & \textbf{0.267} & \textbf{0.851}  \\ \bottomrule
\end{tabular}
\end{center}
\caption{Evaluation results of four methods using \textbf{Qwen2.5-3b-instruct} as the backbone for fine-tuning. } 
\label{table:exp1-qwen}
\vspace*{-0.8cm}
\end{table}

\subsection{Deployment to RAG}
\label{sec:deployment_eval}

\paragraph{Test Dataset.} Unlike previous experiments that partly rely on LLM-generated QAs, this evaluation is conducted on fully human-curated QAs, providing a more realistic and rigorous assessment of model performance. We deploy \texttt{Ext2Gen-R2} in a real RAG environment, retrieving text chunks online from a large corpus in a vector database and prompting LLMs with the target query and the Top-$k$ retrieved chunks. We sample 200 query-answer pairs from each of the three human-annotated RAG benchmarks—Natural Questions (NQ), MS-MARCO, and HotpotQA—totaling 600 examples. For the search corpus, we follow the BEIR benchmark~\cite{thakur2beir}, using 2.7M and 5M text chunks for NQ and HotpotQA, respectively, and adopt the official MS-MARCO setup \cite{bajaj2016ms} with 88M chunks.


\paragraph{Retrieval.} Before generation, we retrieve the Top-$k$ text chunks for each query, varying $k$ in \{10, 20, 30\}. To evaluate the generalization capability of our model, we apply three retrieval methods: \texttt{Naive}, a basic dense retriever using the multilingual-e5-large-instruct model~\cite{wang2024multilingual}, and two advanced retrievers that incorporate query rewriting, namely \texttt{HyDE}\cite{gao2023precise} and \texttt{MuGI}\cite{zhang2024mugi}. The retrieved chunks are added to the \algname{} prompt to extract key sentences and generate the final answer. Importantly, by including retrieval methods enhanced with query rewriting, we aim to verify whether \texttt{Ext2Gen} can still provide substantial gains even when the retrieval results are of higher quality and contain less noise.


\subsubsection{Main Results}
\label{sec:exp2_main}

Figure \ref{fig:deployment} shows a comparison of answer accuracy across three benchmark datasets for two model configurations. The first uses three retrieval methods with the standard Llama3.1-8b-instruct backbone (\texttt{Default}), while the second uses the same backbone trained with \texttt{Ext2Gen-R2}. 

A key observation is that increasing Top-$k$ improves retrieval recall by including more relevant chunks. However, the \texttt{Default} model shows decreased accuracy on NQ and only marginal improvement on HotPotQA. This suggests that as Top-$k$ increases, the precision of retrieval drops, introducing more noise into the input. The \texttt{Default} model struggles to filter out this noisy information and is also sensitive to the position of relevant chunks, which limits its ability to effectively leverage the additional retrieved content.

In contrast, \texttt{Ext2Gen-R2} demonstrates strong robustness, effectively closing the alignment gap where human expectations demand consistent answers despite retrieval-induced noise and uncertain chunk placement, which are the two challenges that standard models struggle to address. By more effectively integrating the retrieved content, \texttt{Ext2Gen-R2} achieves substantial performance gains over \texttt{Default} in real-world RAG settings.

Moreover, while \texttt{Ext2Gen-R2} performs well with the naive retrieval, {advanced query rewriting methods (HyDE, MuGI) exhibit even greater gains when combined with \texttt{Ext2Gen-R2}}, achieving the highest accuracy on NQ and HotPotQA. This highlights that \algname{} not only synergizes with improved retrieval but continues to deliver large performance gains even when retrieval quality is already high. This suggests \texttt{Ext2Gen-R2} addresses limitations that retrieval alone cannot solve, offering a complementary capability on the generation side to better interpret and utilize retrieved content. 

%






\section{Discussion}
\label{sec:discussion}

\paragraph{Generalization to Other Domains with Small LLMs.}
While Ext2Gen is demonstrated primarily on QA-style RAG scenarios, the underlying mechanism---learning to extract relevant evidence before generation---does not inherently depend on domain-specific properties. Instead, it relies on the presence of mixed relevant and irrelevant context, a condition common across domains such as medical IR, scientific literature QA, legal document retrieval, long-context summarization, and enterprise knowledge bases. Although our experiments focus on small LLM backbones (\emph{e.g.}, 3B–8B models), Ext2Gen’s alignment signals target behavioral patterns (evidence selection plus robust generation), rather than capacity-dependent memorization. Therefore, we expect the trained models to generalize to other tasks where (i) retrieval recall brings noisy context, (ii) relevant information may be scattered, or (iii) compact models are required for deployment. However, applying Ext2Gen to other domains may require regenerating domain-specific noisy-retrieval training inputs to ensure proper distributional alignment. This remains a promising direction for future work.

\paragraph{Difference from Methods like Search-R1.}
Ext2Gen shares a similarity with methods such as Search-R1 \cite{jin2025search}, in that both incorporate an intermediate reasoning stage before producing the final answer, thereby reducing hallucinations. Search-R1 achieves this through iterative search, query rewriting, and self-verification loops that actively gather and validate evidence during the reasoning process. In contrast, Ext2Gen operates after the retrieval step, focusing on challenges unique to RAG; namely, uncertain placement of relevant chunks and substantial noise introduced by high-recall retrieval. Rather than performing multi-step exploration, Ext2Gen learns to extract only the sentences that truly ground the answer within a fixed noisy context and then generate the final response, guided by pairwise preference alignment that explicitly rewards robustness under noise. Thus, while both approaches employ structured reasoning, Ext2Gen is distinguished by its specialization for noisy-retrieval environments rather than iterative evidence-seeking search.


\section{Conclusion}
\label{sec:conclusion}



We present \algname{}, an extract-then-generate framework that improves RAG robustness to retrieval noise and chunk misplacement. Leveraging preference-aligned pairwise feedback built from well-curated data, it balances precision and recall in sentence extraction, yielding more reliable answer generation. This approach removes the need for any independent compression module, making its gains especially meaningful. Evaluations on both curated datasets and real-world deployments show consistent improvements over strong baselines, including with high-quality retrieval, underscoring the complementary role of generation-side enhancements.


\section*{Ethical Statement}

Our research focuses on aligning LLMs through a unified extraction and generation framework (\algname{}) to enhance robustness in RAG. Since our work primarily involves model training on publicly available datasets and does not include the collection of sensitive or personally identifiable data, it does not pose direct ethical concerns related to privacy or data security.

\section*{Scientific Artifacts}

The QA pairs and output completions were generated using a diverse set of LLMs to promote variation in response styles and reasoning capabilities. To ensure transparency and reproducibility, we employed publicly available checkpoints for open-source models through the Hugging Face platform, selecting models that span a range of sizes and training paradigms. For proprietary models, including GPT-4o, we utilized OpenAI’s paid API access, which provided access to state-of-the-art generation quality under controlled conditions. A list of the LLMs used is presented in Table \ref{table:llm-source}, enabling full replicability of our experimental setup. The prompt used are also presented in Tables \ref{table:base_prompt}--\ref{table:validity_check}, \ref{table:llm_eval}, and \ref{table:filtering_gpt}.

Regarding the use of generative AI in manuscript preparation, all drafts were written directly by the authors without relying on AI-generated content for core writing. Generative AI tools were employed only for auxiliary editing tasks, such as correcting grammar, refining phrasing, and improving clarity in specific sentences. No AI system was used to generate or draft sections of the manuscript, ensuring that all intellectual contributions reflect the authors' own reasoning and interpretation.


\section*{Acknowledgement}

KAIST was supported by the KISTI graint in 2025 (No.(KISTI) K25L1M1C1), aimed at developing KONI (KISTI Open Neural Intelligence), a large language model specialized in science and technology and by the IITP grant funded by the Korea government (MSIT) (RS-2024-00445087, Enhancing AI Model Reliability Through Domain-Specific Automated Value Alignment Assessment) \& (RS-2025-25464461, AI’s Vision of Harmony: A Fair and Transparent Multimodal Agentic Platform for Conflict Mediation).

\printbibliography

@inproceedings{song2024learning,
  title={Learning to Summarize from LLM-generated Feedback},
  author={Song, Hwanjun and Yun, Taewon and Lee, Yuho and Lee, Gihun and Cai, Jason and Su, Hang},
  booktitle={NAACL},
  year={2025}
}

@inproceedings{yoon2024compact,
  title={CompAct: Compressing Retrieved Documents Actively for Question Answering},
  author={Yoon, Chanwoong and Lee, Taewhoo and Hwang, Hyeon and Jeong, Minbyul and Kang, Jaewoo},
  booktitle={EMNLP},
  year={2024}
}

@article{jin2025search,
  title={Search-r1: Training llms to reason and leverage search engines with reinforcement learning},
  author={Jin, Bowen and Zeng, Hansi and Yue, Zhenrui and Yoon, Jinsung and Arik, Sercan and Wang, Dong and Zamani, Hamed and Han, Jiawei},
  journal={arXiv preprint arXiv:2503.09516},
  year={2025}
}

@inproceedings{hwang2025exit,
  title={Exit: Context-aware extractive compression for enhancing retrieval-augmented generation},
  author={Hwang, Taeho and Cho, Sukmin and Jeong, Soyeong and Song, Hoyun and Han, SeungYoon and Park, Jong C},
  booktitle={ACL},
  year={2025}
}

@inproceedings{wu2025lighter,
  title={Lighter and better: Towards flexible context adaptation for retrieval augmented generation},
  author={Wu, Chenyuan and Shao, Ninglu and Liu, Zheng and Xiao, Shitao and Li, Chaozhuo and Zhang, Chen and Wang, Senzhang and Lian, Defu},
  booktitle={WSDM},
  year={2025}
}

@inproceedings{rau2024context,
  title={Context embeddings for efficient answer generation in rag},
  author={Rau, David and Wang, Shuai and D{\'e}jean, Herv{\'e} and Clinchant, St{\'e}phane},
  booktitle={WSDM},
  year={2025}
}

@inproceedings{choi2025word2passage,
  title={Word2Passage: Word-level Importance Re-weighting for Query Expansion},
  author={Choi, Jeonghwan and Ban, Minjeong and Kim, Minseok and Song, Hwanjun},
  booktitle={ACL},
  year={2025}
}

@inproceedings{fan2024survey,
  title={A survey on rag meeting llms: Towards retrieval-augmented large language models},
  author={Fan, Wenqi and Ding, Yujuan and Ning, Liangbo and Wang, Shijie and Li, Hengyun and Yin, Dawei and Chua, Tat-Seng and Li, Qing},
  booktitle={SIGKDD},
  year={2024}
}

@inproceedings{zhangbertscore,
  title={BERTScore: Evaluating Text Generation with BERT},
  author={Zhang, Tianyi and Kishore, Varsha and Wu, Felix and Weinberger, Kilian Q and Artzi, Yoav},
  booktitle={ICLR},
  year={2020}
}

@inproceedings{hendrycksmeasuring,
  title={Measuring Massive Multitask Language Understanding},
  author={Hendrycks, Dan and Burns, Collin and Basart, Steven and Zou, Andy and Mazeika, Mantas and Song, Dawn and Steinhardt, Jacob},
  booktitle={ICLR},
  year={2021},
}

@article{izacardunsupervised,
  title={Unsupervised Dense Information Retrieval with Contrastive Learning},
  author={Izacard, Gautier and Caron, Mathilde and Hosseini, Lucas and Riedel, Sebastian and Bojanowski, Piotr and Joulin, Armand and Grave, Edouard},
  journal={Transactions on Machine Learning Research},
  year={2021},
}

@article{myrzakhan2024open,
  title={Open-llm-leaderboard: From multi-choice to open-style questions for llms evaluation, benchmark, and arena},
  author={Myrzakhan, Aidar and Bsharat, Sondos Mahmoud and Shen, Zhiqiang},
  journal={arXiv preprint arXiv:2406.07545},
  year={2024}
}

@inproceedings{yang2024rewards,
  title={Rewards-in-Context: Multi-objective Alignment of Foundation Models with Dynamic Preference Adjustment},
  author={Yang, Rui and Pan, Xiaoman and Luo, Feng and Qiu, Shuang and Zhong, Han and Yu, Dong and Chen, Jianshu},
  booktitle={ICML},
  year={2024}
}

@inproceedings{lin2004rouge,
  title={Rouge: A package for automatic evaluation of summaries},
  author={Lin, Chin-Yew},
  booktitle={Text summarization branches out},
  pages={74--81},
  year={2004}
}

@article{gao2023retrieval,
  title={Retrieval-augmented generation for large language models: A survey},
  author={Gao, Yunfan and Xiong, Yun and Gao, Xinyu and Jia, Kangxiang and Pan, Jinliu and Bi, Yuxi and Dai, Yi and Sun, Jiawei and Wang, Haofen},
  journal={arXiv preprint arXiv:2312.10997},
  year={2023}
}

@inproceedings{wang2023query2doc,
  title={Query2{D}oc: Query Expansion with Large Language Models},
  author={Wang, Liang and Yang, Nan and Wei, Furu},
  booktitle={EMNLP},
  year={2023}
}

@inproceedings{asaiself,
  title={Self-RAG: Learning to Retrieve, Generate, and Critique through Self-Reflection},
  author={Asai, Akari and Wu, Zeqiu and Wang, Yizhong and Sil, Avirup and Hajishirzi, Hannaneh},
  booktitle={ICLR},
  year={2024}
}

@inproceedings{zhang2024mugi,
  title={Exploring the Best Practices of Query Expansion with Large Language Models},
  author={Zhang, Le and Wu, Yihong  and Yang, Qian  and Nie, Jian-Yun},
  booktitle={EMNLP},
  year={2024}
}

@inproceedings{li2024you,
  title={Do You Know What You Are Talking About? Characterizing Query-Knowledge Relevance For Reliable Retrieval Augmented Generation},
  author={Li, Zhuohang and Zhang, Jiaxin and Yan, Chao and Das, Kamalika and Kumar, Sricharan and Kantarcioglu, Murat and Malin, Bradley},
  booktitle={EMNLP},
  year={2024}
}

@inproceedings{reddy2024first,
  title={{FIRST}: Faster Improved Listwise Reranking with Single Token Decoding},
  author={Reddy, Revanth and Doo, Jaehyeok and Xu, Yifei and Sultan, Arafat and Swain, Deevya and Sil, Avi and Ji, Heng},
  booktitle={EMNLP},
  year={2024}
}

@article{liu2024lost,
  title={Lost in the middle: How language models use long contexts},
  author={Liu, Nelson F and Lin, Kevin and Hewitt, John and Paranjape, Ashwin and Bevilacqua, Michele and Petroni, Fabio and Liang, Percy},
  journal={Transactions of the Association for Computational Linguistics},
  volume={12},
  pages={157--173},
  year={2024},
}

@inproceedings{yu2024evaluation,
  title={Evaluation of retrieval-augmented generation: A survey},
  author={Yu, Hao and Gan, Aoran and Zhang, Kai and Tong, Shiwei and Liu, Qi and Liu, Zhaofeng},
  booktitle={BigData},
  year={2024},
}

@inproceedings{cuconasu2024power,
  title={The power of noise: Redefining retrieval for rag systems},
  author={Cuconasu, Florin and Trappolini, Giovanni and Siciliano, Federico and Filice, Simone and Campagnano, Cesare and Maarek, Yoelle and Tonellotto, Nicola and Silvestri, Fabrizio},
  booktitle={SIGIR},
  year={2024}
}

@article{rashid2024progressive,
  title={Progressive Query Expansion for Retrieval Over Cost-constrained Data Sources},
  author={Rashid, Muhammad Shihab and Meem, Jannat Ara and Dong, Yue and Hristidis, Vagelis},
  journal={arXiv preprint arXiv:2406.07136},
  year={2024}
}

@inproceedings{ye2024r,
  title={R\^{} 2AG: Incorporating Retrieval Information into Retrieval Augmented Generation},
  author={Ye, Fuda and Li, Shuangyin and Zhang, Yongqi and Chen, Lei},
  booktitle={EMNLP},
  year={2024}
}

@article{yan2024corrective,
  title={Corrective retrieval augmented generation},
  author={Yan, Shi-Qi and Gu, Jia-Chen and Zhu, Yun and Ling, Zhen-Hua},
  journal={arXiv preprint arXiv:2401.15884},
  year={2024}
}

@inproceedings{he2024retrieving,
  title={Retrieving, Rethinking and Revising: The Chain-of-Verification Can Improve Retrieval Augmented Generation},
  author={He, Bolei and Chen, Nuo and He, Xinran and Yan, Lingyong and Wei, Zhenkai and Luo, Jinchang and Ling, Zhen-Hua},
  booktitle={EMNLP},
  year={2024}
}

@article{hwang2024dslr,
  title={{DSLR}: Document Refinement with Sentence-Level Re-ranking and Reconstruction to Enhance Retrieval-Augmented Generation},
  author={Hwang, Taeho and Jeong, Soyeong and Cho, Sukmin and Han, SeungYoon and Park, Jong C},
  journal={arXiv preprint arXiv:2407.03627},
  year={2024}
}

@article{guan2024deliberative,
  title={Deliberative alignment: Reasoning enables safer language models},
  author={Guan, Melody Y and Joglekar, Manas and Wallace, Eric and Jain, Saachi and Barak, Boaz and Heylar, Alec and Dias, Rachel and Vallone, Andrea and Ren, Hongyu and Wei, Jason and others},
  journal={arXiv preprint arXiv:2412.16339},
  year={2024}
}

@article{wei2022chain,
  title={Chain-of-thought prompting elicits reasoning in large language models},
  author={Wei, Jason and Wang, Xuezhi and Schuurmans, Dale and Bosma, Maarten and Xia, Fei and Chi, Ed and Le, Quoc V and Zhou, Denny and others},
  journal={NuerIPS},
  year={2022}
}

@article{chu2023survey,
  title={A survey of chain of thought reasoning: Advances, frontiers and future},
  author={Chu, Zheng and Chen, Jingchang and Chen, Qianglong and Yu, Weijiang and He, Tao and Wang, Haotian and Peng, Weihua and Liu, Ming and Qin, Bing and Liu, Ting},
  journal={arXiv preprint arXiv:2309.15402},
  year={2023}
}

@article{wang2024comprehensive,
  title={A Comprehensive Survey of LLM Alignment Techniques: {RLHF}, {RLAIF}, {PPO}, {DPO} and More},
  author={Wang, Zhichao and Bi, Bin and Pentyala, Shiva Kumar and Ramnath, Kiran and Chaudhuri, Sougata and Mehrotra, Shubham and Mao, Xiang-Bo and Asur, Sitaram and others},
  journal={arXiv preprint arXiv:2407.16216},
  year={2024}
}

@article{zhou2024lima,
  title={Lima: Less is more for alignment},
  author={Zhou, Chunting and Liu, Pengfei and Xu, Puxin and Iyer, Srinivasan and Sun, Jiao and Mao, Yuning and Ma, Xuezhe and Efrat, Avia and Yu, Ping and Yu, Lili and others},
  journal={NeurIPS},
  year={2024}
}

@article{schulman2017proximal,
  title={Proximal policy optimization algorithms},
  author={Schulman, John and Wolski, Filip and Dhariwal, Prafulla and Radford, Alec and Klimov, Oleg},
  journal={arXiv preprint arXiv:1707.06347},
  year={2017}
}

@article{ethayarajh2024kto,
  title={{KTO}: Model alignment as prospect theoretic optimization},
  author={Ethayarajh, Kawin and Xu, Winnie and Muennighoff, Niklas and Jurafsky, Dan and Kiela, Douwe},
  journal={arXiv preprint arXiv:2402.01306},
  year={2024}
}

@article{rafailov2024direct,
  title={Direct preference optimization: Your language model is secretly a reward model},
  author={Rafailov, Rafael and Sharma, Archit and Mitchell, Eric and Manning, Christopher D and Ermon, Stefano and Finn, Chelsea},
  journal={NeurIPS},
  year={2024}
}

@article{robertson2009probabilistic,
  title={The probabilistic relevance framework: BM25 and beyond},
  author={Robertson, Stephen and Zaragoza, Hugo and others},
  journal={Foundations and Trends{\textregistered} in Information Retrieval},
  volume={3},
  number={4},
  pages={333--389},
  year={2009}
}

@article{zhao2024dense,
  title={Dense text retrieval based on pretrained language models: A survey},
  author={Zhao, Wayne Xin and Liu, Jing and Ren, Ruiyang and Wen, Ji-Rong},
  journal={ACM Transactions on Information Systems},
  volume={42},
  number={4},
  pages={1--60},
  year={2024}
}

@inproceedings{yurankrag,
  title={Rank{RAG}: Unifying Context Ranking with Retrieval-Augmented Generation in LLMs},
  author={Yu, Yue and Ping, Wei and Liu, Zihan and Wang, Boxin and You, Jiaxuan and Zhang, Chao and Shoeybi, Mohammad and Catanzaro, Bryan},
  booktitle={NeurIPS}, 
  year={2024}
}

@inproceedings{laban2024summary,
  title={Summary of a Haystack: A Challenge to Long-Context LLMs and RAG Systems},
  author={Laban, Philippe and Fabbri, Alexander Richard and Xiong, Caiming and Wu, Chien-Sheng},
  booktitle={EMNLP},
  year={2024}
}

@inproceedings{islam2024open,
  title={Open-{RAG}: Enhanced Retrieval Augmented Reasoning with Open-Source Large Language Models},
  author={Islam, Shayekh and Rahman, Md Asib and Hossain, KSM Tozammel and Hoque, Enamul and Joty, Shafiq and Parvez, Md Rizwan},
  booktitle={EMNLP},
  year={2024}
}

@inproceedings{xu2023recomp,
  title={{RECOMP}: Improving retrieval-augmented lms with compression and selective augmentation},
  author={Xu, Fangyuan and Shi, Weijia and Choi, Eunsol},
  booktitle={ICLR},
  year={2024}
}

@inproceedings{nallapati2016abstractive,
  title={Abstractive Text Summarization using Sequence-to-sequence RNNs and Beyond},
  author={Nallapati, Ramesh and Zhou, Bowen and dos Santos, Cicero and Gulcehre, Caglar and Xiang, Bing},
  booktitle={SIGNLL},
  year={2016}
}

@inproceedings{huang2021efficient,
  title={Efficient Attentions for Long Document Summarization},
  author={Huang, Luyang and Cao, Shuyang and Parulian, Nikolaus and Ji, Heng and Wang, Lu},
  booktitle={NAACL},
  year={2021}
}

@inproceedings{cohan2018discourse,
  title={A Discourse-Aware Attention Model for Abstractive Summarization of Long Documents},
  author={Cohan, Arman and Dernoncourt, Franck and Kim, Doo Soon and Bui, Trung and Kim, Seokhwan and Chang, Walter and Goharian, Nazli},
  booktitle={NAACL},
  year={2018}
}

@inproceedings{craswell2021ms,
  title={{MS-MARCO}: Benchmarking ranking models in the large-data regime},
  author={Craswell, Nick and Mitra, Bhaskar and Yilmaz, Emine and Campos, Daniel and Lin, Jimmy},
  booktitle={SIGIR},
  year={2021}
}

@inproceedings{li2023synthetic,
  title={Synthetic Data Generation with Large Language Models for Text Classification: Potential and Limitations},
  author={Li, Zhuoyan and Zhu, Hangxiao and Lu, Zhuoran and Yin, Ming},
  booktitle={EMNLP},
  year={2023},
}

@inproceedings{thakur2beir,
  title={{BEIR}: A Heterogeneous Benchmark for Zero-shot Evaluation of Information Retrieval Models},
  author={Thakur, Nandan and Reimers, Nils and R{\"u}ckl{\'e}, Andreas and Srivastava, Abhishek and Gurevych, Iryna},
  booktitle={NeurIPS},
  year={2021},
}

@article{liu2024enhancing,
  title={Enhancing llm safety via constrained direct preference optimization},
  author={Liu, Zixuan and Sun, Xiaolin and Zheng, Zizhan},
  journal={arXiv preprint arXiv:2403.02475},
  year={2024}
}

@article{bajaj2016ms,
title={{MS-MARCO}: A human generated machine reading comprehension dataset},
author={Bajaj, Payal and Campos, Daniel and Craswell, Nick and Deng, Li and Gao, Jianfeng and Liu, Xiaodong and Majumder, Rangan and McNamara, Andrew and Mitra, Bhaskar and Nguyen, Tri and others},
journal={arXiv preprint arXiv:1611.09268},
year={2016}
}

@inproceedings{yang2018hotpotqa,
  title={Hotpot{QA}: A Dataset for Diverse, Explainable Multi-hop Question Answering},
  author={Yang, Zhilin and Qi, Peng and Zhang, Saizheng and Bengio, Yoshua and Cohen, William and Salakhutdinov, Ruslan and Manning, Christopher D},
  booktitle={EMNLP},
  year={2018}
}

@article{wang2024multilingual,
  title={Multilingual e5 text embeddings: A technical report},
  author={Wang, Liang and Yang, Nan and Huang, Xiaolong and Yang, Linjun and Majumder, Rangan and Wei, Furu},
  journal={arXiv preprint arXiv:2402.05672},
  year={2024}
}

@article{das2024under,
  title={Under the surface: Tracking the artifactuality of llm-generated data},
  author={Das, Debarati and De Langis, Karin and Martin-Boyle, Anna and Kim, Jaehyung and Lee, Minhwa and Kim, Zae Myung and Hayati, Shirley Anugrah and Owan, Risako and Hu, Bin and Parkar, Ritik and others},
  journal={arXiv preprint arXiv:2401.14698},
  year={2024}
}

@inproceedings{song2024preference,
  title={Preference ranking optimization for human alignment},
  author={Song, Feifan and Yu, Bowen and Li, Minghao and Yu, Haiyang and Huang, Fei and Li, Yongbin and Wang, Houfeng},
  booktitle={AAAI},
  year={2024}
}

@inproceedings{xu2024pride,
  title={Pride and prejudice: LLM amplifies self-bias in self-refinement},
  author={Xu, Wenda and Zhu, Guanglei and Zhao, Xuandong and Pan, Liangming and Li, Lei and Wang, William},
  booktitle={ACL},
  year={2024}
}

@article{chaudhari2024rlhf,
  title={RLHF Deciphered: A Critical Analysis of Reinforcement Learning from Human Feedback for LLMs},
  author={Chaudhari, Shreyas and Aggarwal, Pranjal and Murahari, Vishvak and Rajpurohit, Tanmay and Kalyan, Ashwin and Narasimhan, Karthik and Deshpande, Ameet and da Silva, Bruno Castro},
  journal={arXiv preprint arXiv:2404.08555},
  year={2024}
}

@article{meng2024simpo,
  title={Sim{PO}: Simple preference optimization with a reference-free reward},
  author={Meng, Yu and Xia, Mengzhou and Chen, Danqi},
  journal={arXiv preprint arXiv:2405.14734},
  year={2024}
}

@inproceedings{gao2023precise,
  title={Precise Zero-Shot Dense Retrieval without Relevance Labels},
  author={Gao, Luyu and Ma, Xueguang and Lin, Jimmy and Callan, Jamie},
  booktitle={ACL},
  year={2023}
}

@article{dettmers2024qlora,
  title={{QLoRA}: Efficient finetuning of quantized llms},
  author={Dettmers, Tim and Pagnoni, Artidoro and Holtzman, Ari and Zettlemoyer, Luke},
  journal={NeurIPS},
  year={2024}
}

@String{Chelsea = "Chelsea" }

\end{document}